\documentclass[twoside,11pt]{article}

\usepackage{jmlr2e}

\usepackage[utf8]{inputenc} % allow utf-8 input
\usepackage[T1]{fontenc}    % use 8-bit T1 fonts
\usepackage{url}            % simple URL typesetting
\usepackage{booktabs}       % professional-quality tables
\usepackage{amsfonts}       % blackboard math symbols
\usepackage{nicefrac}       % compact symbols for 1/2, etc.
\usepackage{microtype}      % microtypography
\usepackage{indentfirst}
\usepackage{float}
\usepackage{graphicx}
\usepackage{subfigure}
\usepackage{listings}
\usepackage{amstext}
\usepackage{amsmath}
\usepackage{amssymb}
\usepackage{xcolor}
\usepackage{dsfont}
\usepackage{algorithm}% http://ctan.org/pkg/algorithms
\usepackage{algpseudocode}%
\newtheorem{condition}{Condition}
\newtheorem{deff}{Definition}
\newtheorem{remm}{Remark}

\ShortHeadings{Empirical Bayes Classification}{Ouyang and Liang}
\firstpageno{1}

\begin{document}

\title{An Empirical Bayes Approach for High Dimensional Classification}

\author{\name Yunbo Ouyang \email youyang4@illinois.edu \\
       \addr Department of Statistics\\
       University of Illinois at Urbana-Champaign\\
       Champaign, IL 61820  USA
       \AND
       \name Feng Liang \email liangf@illinois.edu\\
       \addr Department of Statistics\\
       University of Illinois at Urbana-Champaign\\
       Champaign, IL 61820  USA}

\editor{XXX}

\maketitle

\begin{abstract}%   <- trailing '%' for backward compatibility of .sty file
We propose an empirical Bayes estimator based on Dirichlet process mixture model for estimating the sparse normalized mean difference, which could be directly applied to the high dimensional linear classification. In theory, we build a bridge to connect the estimation error of the mean difference and the misclassification error, also provide sufficient conditions of sub-optimal classifiers and optimal classifiers. In implementation, a variational Bayes algorithm is developed to compute the posterior efficiently and could be parallelized to deal with the ultra-high dimensional case.
\end{abstract}


\section{Introduction}
Nowadays high dimensional classification is ubiquitous in many application areas, such as micro-array data analysis in bioinformatics, document classification in information retrieval, and portfolio analysis in finance.

In this paper, we consider the problem of constructing a linear classifier with  high-dimensional features. Suppose data from class $k$ are generated from a $p$-dimensional multivariate Normal distribution $N_p(\boldsymbol{\mu}_k,\Sigma)$ where $k=1, 2$ and the prior proportions for two classes are $\pi_k$, $k=1,2$ respectively. It is well-known that the optimal classification rule, i.e., the Bayes rule, classifies a new observation $X$ to class $1$ if and only if
\begin{equation} \label{eq:delta}
\delta_{OPT}(X)=(X-\boldsymbol{\mu})^t\Sigma^{-1}\mathbf{d}>\log(\pi_2/\pi_1),
\end{equation}
where $\boldsymbol{\mu}=(\boldsymbol{\mu}_1+\boldsymbol{\mu}_2)/2$ and $d=\boldsymbol{\mu}_1-\boldsymbol{\mu}_2$. For simplicity, we assume both prior proportions $\pi_k$ and sample proportion of two classes are equal, but our theory could be easily extended to the case when two classes have unequal sample size but the ratio is bounded between 0 and 1. Therefore (\ref{eq:delta}) could be simplified as we classifies $X$ to class 1 if and only if
\begin{equation} \label{eq:delta2}
\delta_{OPT}(X)=(X-\boldsymbol{\mu})^t\Sigma^{-1}\mathbf{d}>0.
\end{equation}

Since parameters
$\boldsymbol\theta=(\boldsymbol{\mu}_1,\boldsymbol{\mu}_2,\Sigma)$ are unknown and we are given a set of a random samples $\{ X_{ki}: i=1, \dots, n; k=1,2 \}$, we can estimate those unknown parameters and classify $X$ to class $1$ if \[\hat{\delta}(X)=(X-\hat{\boldsymbol{\mu}})^t\hat{\Sigma}^{-1}\hat{\mathbf{d}}>0,\]
where $\hat{\boldsymbol{\mu}} = \bar{X}_{\cdot \cdot} $ is the overall average of the data, $\hat{d}= \bar{X}_{1 \cdot} -\bar{X}_{2 \cdot}$ is the sample mean difference between the two classes, and $\hat\Sigma = \frac{1}{2 n -2} \sum_{k} \sum_i (X_{k i} - \bar{X}_{k \cdot}) (X_{k i} - \bar{X}_{k \cdot})^t$ is the pooled estimator of the covariance matrix. This is also known as the linear discriminant analysis (LDA).

LDA, however,  doesn't perform well when $p$ is much larger than $n$. \citet{bickel2004some} have shown that when the number of features $p$ grows faster than the sample size, LDA is asymptotically as bad as random guessing  due to the large bias of $\hat{\Sigma}$  in terms of the spectral norm. RDA by \citet{friedman1989regularized}, thresholded covariance matrix estimator by \citet{bickel2008regularized} and Sparse LDA by \citet{shao2011sparse} use regularization to improve the estimation of $\Sigma$ by assuming sparsity on off-diagonal elements of the covariance matrix. \citet{cai2012direct} assume $\Sigma^{-1}\mathbf{d}$ is sparse and proposed LPD based on the sparse estimator of $\Sigma^{-1}\mathbf{d}$.

A seemingly extreme way is to  set all the off-diagonal elements of $\hat{\Sigma}$ to be zero, i.e., ignore the correlation among the $p$ features, and use the following Independence Rule:
\begin{equation}\label{eq:ir}
\hat{\delta}_I(X)=(X-\hat{\boldsymbol{\mu}})^t\hat{\mathbf{D}}^{-1}\hat{\mathbf{d}},
\end{equation}
where $\hat{\mathbf{D}} = \text{diag}(\hat{\Sigma})$.
%$\hat{\mathbf{D}}=\text{diag}\{(S^2_{1j}+S^2_{2j})/2,j=1,2,\cdots,p\}$. $S^2_{kj}$ is the sample variance of the $j$th feature in class $k$, $k=1,2$. Essentially Independence Rule is naive Bayes rule assuming the equal covariance matrix.
Theoretical studies such as \citet{domingos1997optimality} and \citet{bickel2004some} have shown that worst case misclassification error of Independence Rule is well controlled and ignoring the correlation structure of $\Sigma$ doesn't lose much if the correlation matrix is well conditioned.

%There exists different branches of high dimensional classification approaches which could avoid estimating the covariance matrix. Loss-based methods (for example, SVM \citet{vapnik2013nature} and logistic regression) could be refined to add specific penalty terms (such as $L_1$ SVM \citet{zhu20041}, SCAD logistic regression \citet{fan2004nonconcave}). Distance-based approaches like Nearest Shrunken Centroid shrink each class center to the grand mean \citet{guo2007regularized} \citet{tibshirani2003class}.

To achieve good classification performance in a high-dimensional setting, it is not enough to regularize just the covariance matrix. As pointed out by \citet{fan2008high} and \citet{shao2011sparse}, even if we use Independence Rule,  the classification performance of $\hat{\delta}_I$ could still be as bad as random guessing due to the error accumulation through all $p$ dimensions of $\mathbf{d}$. In Theorem 1 of \cite{fan2008high}, essentially using $\hat{\mathbf{d}}$ to estimate $\mathbf{d}$ results in strong condition of signal strength with respect to dimension $p$. If $\mathbf{d}$ is sparse and we use a regularized estimator of $\mathbf{d}$ in Independence Rule, conditions on signal strength should be weakened, which is summarized in Theorem 1 in this paper.

Since estimating sparse $\mathbf{d}$ is equivalent to estimating a sparse high dimensional Gaussian sequence, empirical Bayes methods could be used to get a regularized estimator. First we normalize $\mathbf{d}$: denote the normalized mean difference as $\boldsymbol{\eta}=\sqrt{n/2}\mathbf{D}^{-\frac{1}{2}}\mathbf{d}=(\eta_1,\cdots,\eta_p)^t$, the sample version $\mathbf{y}=\sqrt{n/2}\mathbf{\hat{D}}^{-\frac{1}{2}}\hat{\mathbf{d}}=(Y_1,\cdots,Y_p)^t$. We assume
\begin{align*}
Y_j&\sim N(\eta_j,1),\\
\eta_j &\sim G, j=1,2,\cdots,p,
\end{align*}
where $G$ is unknown prior. $\hat{\boldsymbol{\eta}}$ is an empirical Bayes estimator of $\boldsymbol{\eta}$ based on $\mathbf{y}$.  Independence Rule of this scaled version of $\mathbf{d}$ could be written as
\begin{equation}\label{eq:ir2}
\hat{\delta}_{\hat{\boldsymbol{\eta}}}(X)=(X-\hat{\boldsymbol{\mu}})^t\hat{\mathbf{D}}^{-\frac{1}{2}}\hat{\boldsymbol{\eta}},
\end{equation}
From now on we stick to this scaled version of Independence Rule. Subscript $\hat{\boldsymbol{\eta}}$ indicates this Independence Rule is induced by $\hat{\boldsymbol{\eta}}$.

One branch of empirical Bayes approaches, such as \citet{brown2009nonparametric}, \citet{Jiang:Zhang:2009} and \citet{koenker:mizera:2014},  directly work on marginal likelihood of $Y_j$ . \citet{greenshtein2009application} proposed an empirical Bayes classifier inspired by the empirical Bayes estimator in \citet{brown2009nonparametric} (denoted as EB). Recently, \citet{dicker2016high} also proposed a empirical Bayes classifier based on \citet{koenker:mizera:2014}'s work.
Our goal is to justify that a good empirical Bayes estimator indeed leads to an asymptotically optimal linear classifier, which fills the gap in estimation accuracy and classification performance.

Besides working on marginal likelihood, to take advantage of sparsity structure of $\boldsymbol{\eta}$,  \citet{johnstone2004needles} and \citet{martin:walker:2014} assume a two-group prior $G$ with a positive mass at 0. Recently \citet{Ouyang:Liang:needle} proposed a two-group prior $G$ with the continuous part being a normal mixture and they showed the resulting posterior mean could achieve asymptotical minimax rate established by \citet{donoho:etla:1992}. Therefore applying this estimator (denoted as DP) and its sparse variant (denoted as Sparse DP) could result in a good classification rule.

In this paper, we proposed two empirical Bayes classifiers based on DP estimator and Sparse DP estimator. Compared with \citet{greenshtein2009application}, we establish the theoretical connection between the classification error of (\ref{eq:ir}) and the $L_2$ estimation error of $\hat{\boldsymbol\eta}$ explicitly. In particular, we provide sufficient conditions for a estimator $\hat{\boldsymbol\eta}$ to achieve asymptotical optimal classification accuracy, i.e., the resulting Independence Rule is asymptotically as good as the Bayes rule (\ref{eq:delta2}).

The rest of the paper is organized as follows: in Section 2 we establish the relationship between the estimation error and the classification error. In Section 3, we introduce a variational inference algorithm which returns DP and Sparse DP classifier. We present the empirical results in Section 4 and conclusions and future work in Section 5.

\section{Relationship between the Estimation Error and the Classification Error}
We regard the linear classifier construction as a two step procedure. First we calculated $\mathbf{y}$ and proposed a estimator $\hat{\boldsymbol{\eta}}$ based on $\mathbf{y}$. Second we compute the classifier $\hat{\delta}_{\hat{\boldsymbol\eta}}$: we classifies $X$ to class 1 iff $(X-\hat{\boldsymbol{\mu}})^t\hat{\mathbf{D}}^{-\frac{1}{2}}\hat{\boldsymbol\eta}>0$. We call $\hat{\delta}$ a Independence Rule induced by $\hat{\boldsymbol{\eta}}$.

We use 0-1 loss function to evaluate a linear classifier. Without loss of generality, we assume the new observation $X$ comes from class 1 due to symmetry of our rule. Let $\mathbf{X}$ denote the training data used to construct $\hat{\delta}_{\hat{\boldsymbol{\eta}}}$, the posterior misclassification error of $\hat{\delta}_{\hat{\boldsymbol{\eta}}}$ given parameters $\boldsymbol{\theta}=(\boldsymbol{\mu}_1,\boldsymbol{\mu}_2,\Sigma)$ is
\begin{equation}\label{eq:error}
W(\hat{\delta}_{\hat{\boldsymbol{\eta}}},\boldsymbol{\theta})=P(\hat{\delta}_{\hat{\boldsymbol{\eta}}}(X)\leq 0| \mathbf{X})=\Phi(-\Psi),
\end{equation}
where
\[\Psi=\frac{(\boldsymbol{\mu}_1-\hat{\boldsymbol{\mu}})^t\hat{\mathbf{D}}^{-\frac{1}{2}}\hat{\boldsymbol{\eta}}}{\sqrt{\hat{\boldsymbol{\eta}}^t\hat{\mathbf{D}}^{-\frac{1}{2}}\Sigma \hat{\mathbf{D}}^{-\frac{1}{2}}\hat{\boldsymbol{\eta}}}}.\]
$\Phi(\cdot)$ is standard Normal cumulative distribution function. Let $\mathbf{R}=\mathbf{D}^{-1/2}\Sigma \mathbf{D}^{-1/2}$ be the correlation matrix and $\mathbf{D}=\text{diag}(\Sigma)=(\sigma_{ii})^p_{i=1}$. Consider the following parameter space with three pre-specified constants $\lambda_1,k_1,k_2$ with respect to $p$ ($C_p$ will depend on dimension $p$):
\[\boldsymbol{\Theta}=\{\boldsymbol{\theta}: (\boldsymbol{\mu}_1-\boldsymbol{\mu}_2)^t\mathbf{D}^{-1}(\boldsymbol{\mu}_1-\boldsymbol{\mu}_2)= C_p, \lambda_{\max}(\mathbf{R})\leq \lambda_1, 0<k_1<\sigma_{ii}<k_2,1\leq i\leq p\}.\]
Note that we only bound the largest eigenvalues of $\mathbf{R}$ but the smallest eigenvalue could diverge, leading to diverging condition number of $\mathbf{R}$, which is more general than \citet{bickel2004some}.

Based on $\boldsymbol{\Theta}$, worst case posterior error is defined as
\[W(\hat{\delta}_{\hat{\boldsymbol{\eta}}})=\max_{\boldsymbol{\theta}\in \boldsymbol{\Theta}}W(\hat{\delta}_{\hat{\boldsymbol{\eta}}},\boldsymbol{\theta}).\]
Worst case misclassification error is the expectation of $W(\hat{\delta}_{\hat{\boldsymbol{\eta}}})$ over training data: $\overline{W}(\hat{\delta}_{\hat{\boldsymbol{\eta}}})=\mathds{E}_{\mathbf X}(W(\hat{\delta}_{\hat{\boldsymbol{\eta}}}))$. According to Dominance Convergence Theorem, if $W(\hat{\delta}_{\hat{\boldsymbol{\eta}}})$ converges to a constant $c$, $\overline{W}(\hat{\delta}_{\hat{\boldsymbol{\eta}}})\rightarrow c$ as well. Therefore we only need to study $W(\hat{\delta}_{\hat{\boldsymbol{\eta}}})$.

The misclassification error of the optimal rule $\delta$ given $\boldsymbol\theta \in \boldsymbol{\Theta}$ is $W(\delta,\boldsymbol\theta)=\Phi(-\sqrt{\mathbf{d}^t\Sigma^{-1}\mathbf{d}}/2)\leq \Phi (-\sqrt{C_p}/(2\sqrt{\lambda_1}))$. Therefore
\begin{equation}\label{eq:optimal}
W(\delta_{OPT})=\max_{\boldsymbol{\theta}\in \boldsymbol{\Theta}}W(\delta_{OPT},\boldsymbol\theta)=\Phi (-\sqrt{C_p}/(2\sqrt{\lambda_1}))
\end{equation}

We aim to find a linear classifier such that the performance is as good as the optimal rule asymptotically. The worst case classification error of a good classifier should be approximately equal to the worst case classification error of the optimal Bayes rule. We define the asymptotical optimality and sub-optimality of a classifier in terms of worst case classification error, which is similar with definitions in \citet{shao2011sparse}

\begin{deff}
$\hat{\delta}$ is asymptotically optimal if $W(\hat{\delta})/W(\delta_{OPT})\rightarrow_p 1$.
\end{deff}
\begin{deff}
$\hat{\delta}$ is asymptotically sub-optimal if $W(\hat{\delta})-W(\delta_{OPT})\rightarrow_p 0$.
\end{deff}

$W(\hat{\delta}_{\hat{\boldsymbol{\eta}}})$ is related with the estimation accuracy of $\hat{\boldsymbol\eta}$. Since in many high dimensional classification problems most features are irrelevant, we assume $\boldsymbol{\eta}$ is a $s_n$-sparse vector, where $s_n$ is the number of nonzero elements of $\boldsymbol{\eta}$. Without loss of generality, $S=\{1,2,\cdots,s_n\}$ is the non-zero index set while $S^c=\{s_n+1,s_n+2,\cdots,p\}$ is the zero index set of $\boldsymbol\eta$. $\boldsymbol{\eta}=(\boldsymbol{\eta}_1^t,\mathbf{0}^t)^t$, $\hat{\boldsymbol{\eta}}=(\hat{\boldsymbol{\eta}}^t_1,\hat{\boldsymbol{\eta}}^t_2)^t$, $\hat{\boldsymbol{\eta}}_1$ and $\boldsymbol{\eta}_1$ are $s_n$-dimensional, $\hat{\boldsymbol{\eta}}_2$ is $(p-s_n)$-dimensional.
$L_2$ error to estimate nonzero elements of $\boldsymbol{\eta}$ is $\mathds{E}\|\hat{\boldsymbol{\eta}}_1-\boldsymbol{\eta}_1\|^2=\varepsilon_n$. We assume the following two conditions on $\boldsymbol{\eta}$
\begin{condition}
If $|Y_i|\leq b_n$, then $\hat{\eta}_i=0$.
\end{condition}
\begin{condition}
$\sup_{i\in S^c} \mathds{E}(\hat{\eta}^4_i)< \infty$.
\end{condition}

Remember $\mathds{E}\|\hat{\boldsymbol\eta}-\boldsymbol\eta\|^2=\varepsilon_n+\mathds{E}\|\hat{\boldsymbol{\eta}}_2\|^2$. Condition 1 says $\hat{\eta}_i$ is a thresholded estimator while Condition 2 implies the tail of $\hat{\eta}_i$ isn't too heavy for zero elements. Condition 1 and 2 are used to control $\mathds{E}\|\hat{\boldsymbol{\eta}}_2\|^2$.

To compare the performance of our classifier with the optimal rule, the key quantity involved is weighted squared Euclidean distance $C_p=(\boldsymbol{\mu}_1-\boldsymbol{\mu}_2)^t\mathbf{D}^{-1}(\boldsymbol{\mu}_1-\boldsymbol{\mu}_2)$. Theorem 1 shows $W(\hat{\delta}_{\hat{\boldsymbol{\eta}}})$ is asymptotically close to $W(\delta_{OPT})$ as both $n$ and $p$ are diverging with growth rate constraints among $C_p$, $s_n$ and $\varepsilon_n$.

\begin{theorem}
 Suppose $\hat{\boldsymbol{\eta}}$ satisfies Condition 1-2. $\hat{\delta}_{\hat{\boldsymbol{\eta}}}$ is the classification rule induced by $\hat{\boldsymbol{\eta}}$. We assume $n\rightarrow \infty$, $p\rightarrow \infty$, $\log p=o(n)$, $\sqrt{\log (p-s_n)}=o(b_n)$ and $b_n^2/n\rightarrow 0$. We have
 \[W(\hat{\delta}_{\hat{\boldsymbol{\eta}}})\leq \Phi\left(-\frac{\sqrt{n/8}C_p-O_p(\sqrt{\frac{s_n\varepsilon_n}{n}})-O_p(\sqrt{\varepsilon_nC_p})-O_p(\sqrt{C_p})}{\sqrt{\lambda_1}(\sqrt{nC_p/2}+O_p(\sqrt{\varepsilon_n}))}(1+o_p(1))\right).\]
Furthermore, if $\varepsilon_n/n=o(C_p)$, $\sqrt{s_n\varepsilon_n}/n=o(C_p)$ and $nC_p\rightarrow \infty$  then
\[W(\hat{\delta}_{\hat{\boldsymbol{\eta}}})\leq \Phi\left(-\frac{\sqrt{C_p}}{2\sqrt{\lambda_1}}(1+o_p(1))\right).\]
If $C_p\rightarrow c<\infty$, $\hat{\delta}_{\hat{\boldsymbol{\eta}}}$ is asymptotically optimal; if $C_p\rightarrow \infty$, $\hat{\delta}_{\hat{\boldsymbol{\eta}}}$ is asymptotically sub-optimal.
\end{theorem}

Theorem 1 reveals the relationship between estimation accuracy measure $\varepsilon_n$ and worst case classification error $W(\hat{\delta}_{\hat{\boldsymbol{\eta}}})$ explicitly. Bad performance of Independence Rule in Theorem 1 by \citet{fan2008high}and LDA in Theorem 1-2 by \citet{shao2011sparse} is due to simply using sample mean difference to estimate $\boldsymbol{\eta}$. In those theorems $C_p$ need to dominate $\sqrt{p/n}$ to overcome the estimation loss. However,
if we put sparsity assumptions on $\boldsymbol{\eta}$ and use a thresholded estimator satisfying Condition 1-2, the condition on $C_p$ could be relaxed: $C_p$ should have larger order than $\max(\varepsilon_n,\sqrt{s_n\varepsilon_n})/n$. In \citet{Ouyang:Liang:needle}, $\varepsilon_n$ is bounded by $s_n\log(p/s_n)$. Therefore $C_p$ only needs to grow faster than $(s_n/n)\log(p/s_n)$ to guarantee optimality or sub-optimality of $\hat{\delta}_{\hat{\boldsymbol{\eta}}}$, which weakens conditions.

Meanwhile, we have 2 remarks based on Theorem 1.
\begin{remm}
\normalfont
$p$ could grow exponentially with respect to $n$.
\end{remm}
If $i \in S^c$, $Y_i\sim t_{2n-2}$, centered $t$ distribution with the degree of freedom $2n-2$. Otherwise $Y_i$ follows $t$ distribution with the degree of freedom $2n-2$ and the noncentrality parameter $\sqrt{n/2}(\mu_{2i}-\mu_{1i})$. $b_n$ is chosen to satisfy $b_n\rightarrow \infty$ and $b_n^2/n\rightarrow 0$, which separates relevant features and irrelevant features with the large probability. $b_n$ implicitly determines the relative growth rate between $p$ and $n$. Since $\sqrt{\log (p-s_n)}=o(b_n)$ and $b_n^2/n\rightarrow 0$, $\log p$ could have almost the same growth rate as $n$ in the high dimensional sparse case when $p\gg s_n$.
\begin{remm}
\normalfont
The simple hard thresholding estimator using $\mathbf{y}$ satisfies all the technical conditions.
\end{remm}

These conditions are not strict. We illustrate them using a hard thresholding estimator $\hat{\eta}_i=1_{|Y_i|\leq b_n}Y_i$. Then using Central Limit Theorem we have $\varepsilon_n\sim s_n$. If $s_n/n=o(C_p)$, we could get an asymptotically sub-optimal classifier. If nonzero components of $\mathbf{d}$ are bounded away from 0, $s_n/n=o(C_p)$ is guaranteed. Besides, the fourth moment of central $t$ distribution exists. Therefore the simple hard thresholding estimator satisfies all the technical conditions.

One interesting case is when $C_p\rightarrow \infty$, what conditions we need to put to guarantee optimality. Theorem 2 provides an answer.

\begin{theorem}
Suppose $\hat{\boldsymbol{\eta}}$ satisfies Condition 1-2. $\hat{\delta}_{\hat{\boldsymbol{\eta}}}$ is the classification rule induced by $\hat{\boldsymbol{\eta}}$. If $\varepsilon_nC_p=o(n)$, $\sqrt{\varepsilon_ns_n}=o(n)$, $\log p=o(n)$, $\sqrt{\log (p-s_n)}=o(b_n)$ and $b_n^2/n\rightarrow 0$,  then $\hat{\delta}_{\hat{\boldsymbol\eta}}$ is asymptotic optimal as $n\rightarrow \infty$, $p\rightarrow \infty$ and $C_p\rightarrow \infty$.
\end{theorem}

We need slower growth rate of $C_p$ in Theorem 2. If $C_p$ diverges to infinitely fast, the classification task is relatively easy, but $W(\delta_{OPT})$ converges to 0 faster than the rate of $W(\hat{\delta})$. Therefore our classification rule is not optimal. However, if $C_p$ diverges to infinity slowly, convergence rates of $W(\delta_{OPT})$ and $W(\hat{\delta})$ are comparable. We could prove the ratio of these two converges to 1.

$p$ could still grow exponentially fast with respect to $n$ but we need more constraints about $s_n$. If $\hat{\eta}_i=1_{|Y_i|\leq b_n}Y_i$, we have $\varepsilon_n\sim s_n$. Therefore if $s_nC_p=o(n)$ and $\log p=o(n)$, with the proper choice of $b_n$, our classifier is asymptotically optimal. If $C_p\gg s_n$, we have $s^2_n=o(n)$. Therefore $s_n$ must grow slower than $\sqrt{n}$, since we need enough data to estimate $s_n$ nonzero elements accurately to guarantee we have small estimation error.

Any good estimator of the sparse mean difference should have small estimation error leading to small growth rate of $\varepsilon_n$. Our previous work have shown clustering algorithm based estimators have the estimation error $\mathds{E}\|\hat{\boldsymbol\eta}-\boldsymbol\eta\|^2\sim s_n\log(p/s_n)$. Hence, our proposed Dirichlet process mixture method based estimator as a special example, is asymptotically optimal in the minimax criteria. We can gain estimation accuracy in the first step, resulting in the better classification performance.

\section{Dirichlet Process Mixture Based Linear Classifier}
\subsection{Dirichlet Process Prior}
Given $\mathbf{y}=\sqrt{n/2}\mathbf{\hat{D}}^{-\frac{1}{2}}\hat{\mathbf{d}}$, in this section we build an empirical Bayes model with Dirichlet process prior to estimate $\boldsymbol{\eta}=\sqrt{n/2}\mathbf{D}^{-\frac{1}{2}}\mathbf{d}$.  We assume $Y_j\sim N(\eta_j,1)$ and  $\eta_j\sim G$, where $G$ is prior unknown.  Since most of $\eta_j$'s are sparse, $Y_j$'s will concentrate around zero, forming a large cluster at 0 and several other clusters far away from 0. Dirichlet process mixture model is one of Bayesian tools to capture clustering behaviors (see \citet{lo1984class}). We build a hierarchical Bayes model and assume $G\sim DP(\alpha, G_0)$, where $\alpha$ is the concentration parameter and $G_0$ is the base measure.

An important formulation of Dirichlet process is stick breaking process proposed by \citet{sethuraman1994constructive}. We represent the random distribution function $G$ as $\sum^\infty_{t=1}\pi_t\delta_{\eta_t}(\cdot)$, where $\eta_t$ is drawn i.i.d. from the base measure $G_0$, while $\pi_t=V_t\prod^{t-1}_{i=1}(1-V_i)$. $V_i$ is drawn i.i.d. from $\text{Beta}(1,\alpha)$. To guarantee $G$ has a positive mass at 0, we model $G_0$ as Normal distribution with a point mass at 0, that is, $G_0=w\delta_0+(1-w)N(0,\sigma^2)$, where $w$ and $\sigma^2$ are 2 pre-specified parameters. $\delta_0$ is a dirac function at 0.

\subsection{Variational Inference}
To calculate the posterior for stick breaking process representation of Dirichlet process, a common technique is to pre-specify $T$ as the upper bound of the number of clusters. Then we have the following truncated version of stick breaking process using $G_0=w\delta_0+(1-w)N(0,\sigma^2)$ as the base measure:

\begin{align}
&V_t|\alpha \sim \text{Beta}(1,\alpha), t=1,2,\cdots, T-1, V_T=1;\\
&\xi_t\sim \text{Ber}(w), t=1,2,\cdots, T;\\
&\eta^\ast_t|\xi_t \sim \left\{\begin{aligned}
&\delta_0 & \xi_t=1\\
&\text{N}(0,\sigma^2) & \xi_t=0\\
\end{aligned}\right.; t=1,2,\cdots,T;\\
&\pi_t =V_t\prod^{t-1}_{j=1}(1-V_j), t=1,2,\cdots, T-1, \pi_T =\prod^{T}_{j=1}(1-V_j);\\
&Z_k|\{V_1, V_2, \cdots, V_{T-1}\} \sim \text{Multinomial}(\boldsymbol{\pi});\\
&Y_k |Z_k \sim \text{N}(\eta^\ast_{Z_k},1), k=1,2,\cdots, p.
\end{align}

The observed data are $\mathbf{y}$ and the parameters are $\mathbf{Z}_{1\times p},\mathbf{V}_{1\times (T-1)},\boldsymbol{\eta}^\ast_{1\times T},\boldsymbol{\xi}_{1\times T}$. $\boldsymbol{\eta}^\ast=(\eta_1^\ast,\cdots,\eta_T^\ast)$ contains all unique values of $\boldsymbol\eta=(\eta^\ast_{Z_k})^p_{k=1}$.

The number of parameters we estimate is $O(p)$, making MCMC converging very slowly. Instead, we use a variational Bayes algorithm to compute the posterior distribution which has the similar performance as traditional MCMC algorithms. \citet{blei2006variational} propose variational inference algorithms for Dirichlet process mixture model for the exponential family base measure $G_0$ . Although Normal distribution with a positive mass at 0 doesn't belong to exponential family, we could use the similar framework to derive our own variational Bayes algorithm.

We assume the following fully factorized variational distribution:
\[q(\mathbf{Z},\mathbf{V},\boldsymbol{\eta}^\ast,\boldsymbol{\xi})=q_{\mathbf{p},\mathbf{m},\boldsymbol{\tau}}(\boldsymbol{\eta}^\ast,\boldsymbol{\xi})q_{\boldsymbol{\gamma}_1,\boldsymbol{\gamma}_2}(\mathbf{V})q_{\boldsymbol{\Phi}}(\mathbf{Z}).\]
Shown in the Appendix, we've proved
\begin{itemize}
\item
$q_{\mathbf{p},\mathbf{m},\boldsymbol{\tau}}(\boldsymbol{\eta}^\ast,\boldsymbol{\xi})=\prod^T_{t=1}q_{p_t,m_t,\tau_t}(\eta^\ast_t,\xi_t)$,
where $\mathbf{p}=(p_1,p_2,\cdots,p_T)$, $\mathbf{m}=(m_1,m_2,\cdots,m_T)$, $\boldsymbol{\tau}=(\tau_1,\tau_2,\cdots,\tau_T)$, and $q_{p_t,m_t,\tau_t}(\eta^\ast_t,\xi_t)=p_t1_{\xi_t=1}\delta_0+(1-p_t)1_{\xi_t=0}q_{m_t,\tau_t}(\eta^\ast_t)$, where $q_{m_t,\tau_t}(\eta^\ast_t)$ is Normal density with mean $m_t$ and variance $\tau_t^2$.
\item
$q_{\boldsymbol{\gamma}_1,\boldsymbol{\gamma}_2}(\mathbf{V})=\prod^{T-1}_{t=1}q_{\gamma_{1t},\gamma_{2t}}(V_t)$, where
$\boldsymbol{\gamma}_1=(\gamma_{11},\gamma_{12},\cdots,\gamma_{1(T-1)})$, $\boldsymbol{\gamma}_2=(\gamma_{21},\gamma_{22},\cdots,\\ \gamma_{2(T-1)})$,
$q_{\gamma_{1t},\gamma_{2t}}(V_t)$ is Beta Distribution with parameters $(\gamma_{1t},\gamma_{2t})$.
\item
$q_{\boldsymbol{\Phi}}(\mathbf{Z})=\prod^p_{k=1}q_{\boldsymbol{\phi}_k}(Z_k)$;
where  $\mathbf{Z}=(Z_1,Z_2,\cdots,Z_p)$,$\boldsymbol{\Phi}=(\boldsymbol{\phi}_1,\boldsymbol{\phi}_2,\cdots, \boldsymbol{\phi}_p)$,$\boldsymbol{\phi}_k=(\phi_{k,1},\phi_{k,2},\cdots, \phi_{k,T})$
, $\phi_{k,t}=q(Z_k=t)$, $q_{\boldsymbol{\phi}_k}(Z_k)$ is Multinomial distribution with parameters $\boldsymbol{\phi}_k$.
\end{itemize}

The algorithm is summarized in Algorithm 1. Via iterating these steps we could update the variational parameters. After convergence of $\boldsymbol{\Phi}$, $\mathbf{p}$, $\mathbf{m}$, $\boldsymbol{\tau}$, $\boldsymbol{\gamma}_1$ and $\boldsymbol{\gamma}_2$, we get an approximation of the posterior by plugging in these estimated parameters. The parameters we are interested in are $\boldsymbol{\Phi}$,$\mathbf{p}$ and $\mathbf{m}$. (In the algorithm $\|\cdot\|_{\infty,\infty}$ means the element-wise maximum absolute value; $\text{logit}(x)=(1+\exp(-x))^{-1}$.)

\begin{algorithm}[htbp]
\begin{algorithmic}
\State \textbf{input }{$\mathbf{y},\alpha,\sigma,w,T$}
\State \textbf{initialize} $\boldsymbol\Phi^{(1)}$ and $\boldsymbol\Phi^{(0)}$;
\While{$\|\boldsymbol\Phi^{(1)}-\boldsymbol\Phi^{(0)}\|_{\infty,\infty}>\epsilon$}
\State  $m_t\leftarrow\frac{\sigma^2\cdot \sum^p_{n=1}\phi^{(0)}_{k,t}Y_k}{\sigma^2\cdot \sum^p_{k=1}\phi^{(0)}_{k,t}+1}, t=1,2,\cdots, T$;
\State   $\tau_t^2 \leftarrow\frac{\sigma^2}{\sigma^2\cdot \sum^p_{k=1}\phi^{(0)}_{k,t}+1}, t=1,2,\cdots,T$;
\State   $p_t\leftarrow\text{logit}^{-1}(\log(w)-\log(1-w)+\log(\sigma^2\cdot \sum^p_{k=1}\phi^{(0)}_{k,t}+1)/2-\frac{\sigma^2\cdot (\sum^p_{k=1}\phi^{(0)}_{k,t}Y_k)^2}{2(\sigma^2\cdot \sum^p_{k=1}\phi^{(0)}_{k,t}+1)}),t=1,2,\cdots,T$;
\State   $\gamma_{t,1}\leftarrow 1+\sum^{p}_{k=1}\phi^{(0)}_{k,t}, t=1,2,\cdots, T-1$;
 \State $\gamma_{t,2}\leftarrow\alpha+\sum^{p}_{k=1}\sum^{T}_{j=t+1} \phi^{(0)}_{k,t}, t=1,2,\cdots, T-1$;
\State   $S_{k,t}\leftarrow\mathds{E}_{q}\log V_t+ \sum^{t-1}_{i=1}\mathds{E}_{q}\log (1-V_t)+ (1-p_t)m_{t}Y_k-\frac{1}{2}(1-p_t)(m_{t}^2+\tau^2_t),t=1,2,\cdots, T, k=1,2,\cdots, p$;
 \State  $\phi^{(1)}_{k,t}\propto \exp(S_{k,t}), t=1,2,\cdots, T, k=1,2,\cdots, p$;
 \EndWhile
 \State \textbf{output } {$\mathbf{p},\mathbf{m},\boldsymbol{\Phi}$}
\end{algorithmic}
 \vspace{3mm}
 \caption{Variational Bayes Algorithm for Dirichlet process mixture model with $G_0$}
\end{algorithm}

\subsection{Constructing Linear Classifier}
Given approximate posterior estimates $\hat{\mathbf{p}}$, $\hat{\mathbf{m}}$, $\hat{\boldsymbol\Phi}$, we get a MAP (maximum a posterior) estimator of $G$. $m_t$ is the nonzero center and $p_t$ is the probability mass of zero of component indexed by $t$. Each entry $\phi_{kt}$ of $\hat{\boldsymbol{\Phi}}$ is the posterior probability of $Z_k$ belonging to the cluster $t$. Furthermore, approximate posterior distribution of $\eta^\ast_{Z_k}$ is $(\sum^{T}_{t=1}\hat{\phi}_{kt}\hat{p}_t)\delta_0(\cdot)+\sum^T_{t=1}\hat{\phi}_{kt}(1-\hat{p}_t)\delta_{\hat{m}_t}(\cdot)$.
The most probable posterior assignment of $\eta^\ast_{Z_k}$ based on above posterior is denoted as $\hat{\eta}^\ast_{Z_k}$. MAP estimate of cluster weights including zero clusters is $\tilde{w}_t=\#\{k:\hat{\eta}^\ast_{Z_k}=\hat{m}_t\}/p$ and $\tilde{w}_0=\#\{k:\hat{\eta}^\ast_{Z_k}=0\}/p$. Then the estimated prior is $\hat{G}= \tilde{w}_0\cdot\delta_0(\cdot)+\sum^{T}_{t=1}\tilde{w}_t\delta_{\hat{m}_t}(\cdot)$.

Based on $\hat{G}$, the posterior distribution of $\eta_k$ given $Y_k$ is
\[\frac{\tilde{w}_0\exp(-\frac{Y_k^2}{2})\delta_0+\sum^T_{t=1}\tilde{w}_t\exp(-\frac{(Y_k-\hat{m}_t)^2}{2})\delta_{\hat{m}_t}}{\tilde{w}_0\exp(-\frac{Y_k^2}{2})+\sum^T_{t=1}\tilde{w}_t\exp(-\frac{(Y_k-\hat{m}_t)^2}{2})}\equiv \hat{w}_{k0}\cdot\delta_0(\cdot)+\sum^{T}_{t=1}\hat{w}_{kt}\delta_{\hat{m}_t}(\cdot),\]
where $\hat{w}_{kt}$ is the posterior weight. We propose a posterior mean estimator $\hat{\eta}^{\text{DP}}_{k}=\sum^{T}_{t=1}\hat{w}_{kt}\hat{m}_t$. The linear classification rule induced by $\hat{\boldsymbol{\eta}}_{\text{DP}}=(\hat{\eta}^{\text{DP}}_{1},\cdots,\hat{\eta}^{\text{DP}}_{p})^t$ is: we classifies $X$ to class 1 iff
\[\hat{\delta}_{\text{DP}}(X)=(X-\hat{\boldsymbol{\mu}})^t\hat{\mathbf{D}}^{-\frac{1}{2}}\hat{\boldsymbol\eta}_{\text{DP}}>0.\]
We refer to $\hat{\boldsymbol\eta}_{DP}$ as DP estimator and the corresponding classifier as Dirichlet process linear classifier (DP linear classifier).

Additional sparsity could be introduced to DP estimator. Since we use the posterior mean as the estimator, the resulting $\hat{\boldsymbol{\eta}}$ is a shrinkage estimator of the true mean difference but not necessarily sparse. To have better performance in the high dimensional extremely sparse case, we revise the original DP estimator via thresholding posterior probability at 0: if the posterior weight $\hat{w}_{k0}>\kappa$, $\hat{\eta}^{\text{SDP}}_k=0$, otherwise $\hat{\eta}^{\text{SDP}}_k=\hat{\eta}^{\text{DP}}_{k}$, where $\kappa$ is a tuning parameter which could be determined by cross validation. In all the simulation studies we fix $\kappa=0.5$ since choosing the threshold at 0.5 is equivalent to getting a MAP estimator of index set of zeros. We refer to $\hat{\boldsymbol\eta}_{\text{SDP}}=(\hat{\eta}^{\text{SDP}}_{1},\cdots,\hat{\eta}^{\text{SDP}}_{p})^t$ as Sparse DP estimator and the resulting linear classifier $\hat{\delta}_{\text{SDP}}(X)=(X-\hat{\boldsymbol{\mu}})^t\hat{\mathbf{D}}^{-\frac{1}{2}}\hat{\boldsymbol\eta}_{\text{SDP}}$ as Sparse DP linear classifier. Sparse DP estimator is a thresholded estimator whereas DP estimator isn't. Therefore Sparse DP estimator satisfies Condition 1. Sparse DP estimator could eliminate noise of irrelevant features completely to enhance classification performance.

One practical issue of both DP and sparse DP estimator, is when $\boldsymbol\eta$ is extremely sparse, we might end up with a MAP estimator $\hat{G}=\delta_0$ occasionally. This is due to the ``Rich gets richer'' property of Dirichlet Process prior. A remedy in this extreme case is to randomly equally divide all $p$ sample mean differences into $I$ folds. For each fold of data we use Dirichlet process mixture model to estimate the discrete prior $\hat{G}_i$. Then we average all the discrete priors to get a overall estimate $\hat{G}=\sum^I_{i=1}\hat{G}_i/I$. For DP estimator and Sparse DP estimator we both use this refinement to estimate $\boldsymbol\eta$. The rationale behind this ``batch'' processing idea is when we divide elements of a high-dimensional vector into several batches, not only do the relatively large elements pop out because the maximum of the noise decreases as the sample size is smaller, but also the probability of all $\hat{G}_i$s equal to 0 is extremely small. The chance of detecting signals is increased. This refinement naturally leads to a parallelized variational Bayes algorithm: we could parallelize our algorithm for every batch and then average the estimated prior.

\section{Empirical Studies}
In this section, we conducted three simulation studies and applied our method to one real data example. The corresponding R package VBDP is available in \url{https://github.com/yunboouyang/VBDP}, which includes code to estimate sparse Gaussian sequence and code to construct DP and Sparse DP classifiers. Real data example is also included in this package. The source code and simulation results are available in \url{https://github.com/yunboouyang/EBclassifier}. Parameter specification is also summarized in the source code. 

We also include a column ``Hard Thresh DP'' for comparison: Hard Threshold DP classifier uses the same threshold as Sparse DP classifier, but instead of using posterior mean, Hard Threshold DP classifier just uses sample mean difference to estimate $\mathbf{d}$ if the posterior probability at 0 is below threshold. $\varepsilon_n$ is large for Hard Threshold DP classifier but small for DP classifier and Sparse DP classifier because only the last two methods apply shrinkage. The purpose to include Hard Threshold DP classifier is to demonstrate the influence of $\varepsilon_n$ on classification error $W(\hat\delta)$. If $\varepsilon_n$ is large, $W(\hat\delta)$ should be large. We don't recommend to use Hard Threshold DP classifier in practice.

\subsection{Simulation Studies}

We conducted three simulation studies. The first two  are the same  in \citet{greenshtein2009application}. In the third simulation study we compare our methods with \citet{fan2008high} in the same setting.

{\bf Simulation Study 1.}  We assume $\Sigma$ has only diagonal elements. Without loss of generality, we set $\boldsymbol{\mu}_2=\mathbf{0}$ and $\boldsymbol{\mu}_1\neq \mathbf{0}$. We use $(\Delta,l)$ to denote different configurations of $\boldsymbol{\mu}_1$: the first $l$ coordinates in $\boldsymbol{\mu}_1$ are all valued $\Delta$ while the remaining entries are all 0 or sampled from $N(0,0.1^2)$. In the first simulation study, $\Sigma=s^2\mathbf{I}_{p}$, where $s^2=25/2$ and $p=10^4$. The sample size of each class is $n=25$. We compute the theoretical misclassification rate using the true mean and true covariance matrix. We repeat our procedures 100 times and the average theoretical misclassification rates are reported in Table 1 and Table 2 corresponding to different $\boldsymbol{\mu}_1$. Bold case in all tables indicates the lowest misclassification rate across each row.

\begin{table}[H]
\centering
\begin{tabular}{cccccccc}
\hline
$(\Delta,l)$ & Hard Thresh DP & Sparse DP & DP  & EB & IR & FAIR & glmnet  \\
\hline
(1,2000) &0.0046 & 0.0003& \textbf{0.0002}  & 0.0004 & 0.0049 & 0.1211 & 0.4280\\
(1,1000) &0.0874& 0.0454& \textbf{0.0283} & 0.0428 & 0.0885 & 0.2393 & 0.4500\\
(1,500)  &0.2423& 0.2036& \textbf{0.1858}  & 0.2015 & 0.2435 & 0.3423 & 0.4750\\
(1.5,300) &0.1756& 0.1303& \textbf{0.1059} & 0.1160 & 0.1767 & 0.2222 & 0.4146\\
(2,200) &0.1362& 0.0540& \textbf{0.0412} & 0.0518 & 0.1372 & 0.1039 & 0.3046\\
(2.5,100) &0.1937& 0.0449& \textbf{0.0422} & 0.0585 & 0.1947 & 0.0852 & 0.2126\\
(3,50) &0.2652& \textbf{0.0470}& 0.0677 & 0.0772 & 0.2665 & 0.0982 & 0.1498\\
(3.5,50) &0.1957&\textbf{0.0066}& 0.0175 & 0.0152 & 0.1965 & 0.0229 & 0.0655\\
(4,40) &0.1883&\textbf{0.0023} & 0.0059  & 0.0072 & 0.1901 & 0.0101 & 0.0332\\
\hline
\end{tabular}
\caption{Misclassification error rates, $p=10^4$, $p-l$ entries are 0}
\end{table}

\begin{table}[H]
\centering
\begin{tabular}{cccccccc}
\hline
$(\Delta,l)$ & Hard Thresh DP &Sparse DP & DP  & EB & IR & FAIR & glmnet  \\
\hline
(1,2000) &0.0035& 0.0002& \textbf{0.0001}  & 0.0003 & 0.0038 & 0.1128 & 0.4216\\
(1,1000) &0.0699&0.0395 & \textbf{0.0241}  & 0.0352 & 0.0710 & 0.2311 & 0.4551\\
(1,500)  &0.2046&0.1948 & \textbf{0.1686}  & 0.1751 & 0.2063 & 0.3280 & 0.4783\\
(1.5,300) &0.1450& 0.1173& \textbf{0.0976}  & 0.0996 & 0.1465 & 0.2075 & 0.4190\\
(2,200) &0.1102&0.0470 & \textbf{0.0372}  & 0.0431 & 0.1113 & 0.1011 & 0.3158\\
(2.5,100)& 0.1583& \textbf{0.0392}& 0.0415 & 0.0488 & 0.1595 & 0.0815 & 0.1945\\
(3,50) &0.2248&\textbf{0.0444} & 0.0674  & 0.0687 & 0.2265 & 0.0969 & 0.1692\\
(3.5,50) &0.1637&\textbf{0.0065} & 0.0119  & 0.0146 & 0.1655 & 0.0226 & 0.0640\\
(4,40) &0.1539& \textbf{0.0019} & 0.0056  & 0.0057 & 0.1551 & 0.0088 & 0.0324\\
\hline
\end{tabular}
\caption{Misclassification error rates, $p=10^4$, $p-l$ entries are generated from $\text{N}(0,0.1^2)$}
\end{table}

Table 1 and Table 2 compare DP and Sparse DP linear classifier with several existing methods: Empirical Bayes classifier (EB) by \citet{greenshtein2009application}, Independence Rule (IR) by \citet{bickel2004some}, Feature Annealed Independence Rule (FAIR) by \citet{fan2008high} and logistic regression with lasso using R package \textsf{glmnet} (denoted as glmnet).

DP and Sparse DP methods dominate other methods in the diagonal covariance matrix case whether the mean difference is sparse or not. If the mean difference vector is extremely sparse while the signal is strong, Sparse DP classifier outperforms DP classifier. In the relatively dense signal case, DP classifier outperforms sparse DP classifier. Overall DP and sparse DP estimators could improve estimation accuracy of the nonzero true mean difference while ruling out irrelevant features. Hard Thresh DP classifier has similar performance as IR, indicating if estimation error is not well controlled, classification accuracy could not be guaranteed.

{\bf Simulation Study 2.}  We consider AR(1) covariance structure of $\Sigma=s^2\mathbf{R}$, where $s^2=25/2$. That is, the correlation satisfies $R_{ij}=\text{Corr}(X_{kmi},X_{kmj})=\rho^{|i-j|}, k=1,2, 1\leq m\leq n, 1\leq i,j\leq p$. $p=10^4$. Sample size of each class is $n=25$. We consider 3 different configurations of $\boldsymbol{\mu}_1$ in this simulation study. The simulation results are shown in Table 3 to Table 5 based on 100 repetitions to compare theoretical misclassification rates.

\begin{table}[H]
\centering
\begin{tabular}{cccccccc}
\hline
$\rho$ &Hard Thresh DP &Sparse DP &DP & EB  & IR & FAIR & glmnet   \\
\hline
0.3 &0.0089 &0.0031&\textbf{0.0021}  & 0.0022 & 0.0092 & 0.1276 & 0.4325 \\
0.5 &0.0235 &0.0135&0.0105  & \textbf{0.0096}  & 0.0237 & 0.1393 & 0.4340 \\
0.7 & 0.0714&0.0539&0.0468  & \textbf{0.0430}  & 0.0712 & 0.1800 & 0.4437 \\
0.9 &0.2079 &0.1929&0.1867 & \textbf{0.1758}  & 0.2073 & 0.2702 & 0.4586 \\
\hline
\end{tabular}
\caption{Misclassification error rates, $p=10^4$, 2000 entries are 1 for $\boldsymbol{\mu}_1$. Other entries are generated from $\text{N}(0,0.1^2)$}
\end{table}

\begin{table}[H]
\centering
\begin{tabular}{cccccccc}
\hline
$\rho$ &Hard Thresh DP & Sparse DP & DP & EB & IR & FAIR & glmnet   \\
\hline
0.3 &0.0237 &0.0081 &\textbf{0.0056}  & 0.0068  & 0.0243 & 0.0546 & 0.2315 \\
0.5 &0.0481&0.0233 &\textbf{0.0183} & 0.0203 & 0.0483 & 0.0699 & 0.2792 \\
0.7 &0.1036&0.0686 &\textbf{0.0612} & 0.0619  & 0.1033 & 0.1095 & 0.2978 \\
0.9 &0.2472&0.2111  &0.2054  & \textbf{0.2024}  & 0.2466 & 0.2308 & 0.3603 \\
\hline
\end{tabular}
\caption{Misclassification error rates, $p=10^4$, 1000 entries are 1 for $\boldsymbol{\mu}_1$. 100 entries are 2.5. Other entries are generated from $\text{N}(0,0.1^2)$}
\end{table}

\begin{table}[H]
\centering
\begin{tabular}{cccccccc}
\hline
$\rho$ &Hard Thresh DP& Sparse DP & DP  & EB  & IR & FAIR & glmnet   \\
\hline
0.3 &0.0233 &0.0037&\textbf{0.0032}  & 0.0038  & 0.0238 & 0.0226 & 0.0913 \\
0.5 &0.0478 &0.0139&\textbf{0.0129}  & 0.0138  & 0.0475 & 0.0374 & 0.1290 \\
0.7 &0.1069 &0.0508&\textbf{0.0493}  &0.0502  & 0.1069 & 0.0801 & 0.1747 \\
0.9 &0.2445 &\textbf{0.1827}& 0.1871 & 0.1834  & 0.2441 & 0.2067 & 0.2971 \\
\hline
\end{tabular}
\caption{Misclassification error rates, $p=10^4$, 1000 entries are 1 for $\boldsymbol{\mu}_1$. 50 entries are 3.5. Other entries are generated from $\text{N}(0,0.1^2)$}
\end{table}

DP family and EB are among the best methods in this AR(1) correlation structure except Hard Thresh DP. If the correlation is severe and there aren't very large mean difference, EB has better performance. If the correlation isn't extremely severe or there are some large mean difference, DP classifier has better performance. As $\rho$ gets larger, the misclassification rate keeps increasing for each method, Sparse DP classifier and DP classifier is still considered as 2 relatively good classifiers since we only have very few data points.

{\bf Simulation Study 3.} We consider the same setting used in \citet{fan2008high}. The error vector is no longer normal and the covariance matrix has a group structure. All features are divided into 3 groups. Within each group, features share one unobservable common factor with different factor loadings. In addition, there is an unobservable common factor among all the features across 3 groups. $p=4500$ and $n=30$. To construct the error vector, let $Z_{ij}$ be a sequence of independent standard normal random variables, and $\chi^2_{ij}$ be a sequence of independent random variables of the same distribution as $(\chi^2_6-6)/\sqrt{12}$. Let $a_j$ and $b_j$ be factor loading coefficients. Then the error vector
 for each class is defined as
\[\epsilon_{ij}=\frac{Z_{ij}+a_{1j}\chi_{1i}+a_{2j}\chi_{2i}+a_{3j}\chi_{3i}+b_j\chi_{4i}}{\sqrt{1+a^2_{1j}+a^2_{2j}+a^2_{3j}+b_j^2}},i=1,2,\cdots,30,
j=1,2,\cdots, 4500,\]
where $a_{ij}=0$ except that $a_{1j}=a_j$ for $j=1,\cdots, 1500$, $a_{2j}=a_j$ for $j=1501,\cdots, 3000$, and $a_{3j}=a_j$ for $j=3001,\cdots,4500$. Therefore $\mathds{E}(\epsilon_{ij})=0$ and $\text{Var}(\epsilon_{ij})=1$, and in general within-group correlation is greater than the between-group correlation. The factor loadings $a_j$ and $b_j$ are independently generated from uniform distributions $U(0,0.4)$ and $U(0,0.2)$. The mean vector $\boldsymbol{\mu}_1$ is taken from a realization of the mixture of a point mass at 0 and a double exponential distribution:
$(1-c)\delta_0+\frac{1}{2}c\exp(-2|x|)$,
where $c=0.02$. $\boldsymbol{\mu}_2=\mathbf{0}$. There are only very few features with signal levels exceeding 1 standard deviation of the noise. We apply Hard Thresh DP, Sparse DP, FAIR and glmnet to 400 test samples generated from the same process and calculate the average error rate. We also compare these methods to oracle procedure, which we know the location of each nonzero element in $\boldsymbol{\mu}_2$ vector and use these nonzero elements to construct Independence Rule based classifier.  We have 100 repetitions. The boxplot and scatter plot of misclassification error of these 4 methods are summarized in Figure 1 and the average error is summarized in Table 6.
\begin{figure}
\centering
\includegraphics[width=0.7\textwidth]{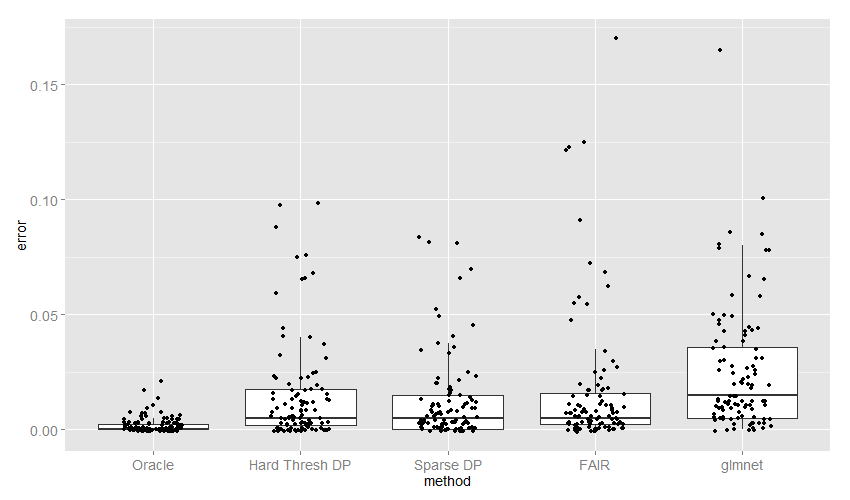}
\caption{Box plot of classification errors of 4 methods}
\end{figure}

\begin{table}[H]
\centering
\begin{tabular}{ccccc}
\hline
Oracle &Hard Thresh DP & Sparse DP & FAIR & glmnet  \\
\hline
0.0021 & 0.0150 & 0.0126 & 0.0168 & 0.0252\\
\hline
\end{tabular}
\caption{Average Misclassification Rate for Simulation Study 3}
\end{table}

Both Hard Thresh DP and Sparse DP classifier are better than FAIR and outperforms the logistic regression with Lasso. Even though on average oracle procedure's misclassification error is smaller than that of DP family classifiers, we could conclude from the plot the misclassification error of majority of 100 trials for Sparse DP classifier is comparable to the misclassification error of the oracle procedure. Sparse DP classifier still has very good performance except some extreme cases.

\subsection{Real Data Example}
We consider a leukemia data set which was first analyzed by \cite{golub1999molecular} and widely used in statistics literature. The data set can be downloaded in \textsf{http://www.broad.mit.edu/cgi-bin/cancer/datasets.cgi}. There are 7129 genes and 72 samples generated from two classes, ALL (acute lymphocytic leukemia) and AML (acute mylogenous leukemia). Among the 72 samples, the training data set has 38 (27 data points in ALL and 11 data points in AML) and the test data set has 34 (20 in ALL and 14 in AML). We compared DP and sparse DP classifier with IR, EB and FAIR, which was summarized in Table 7. For DP and sparse DP classifier, we set $\alpha=1$, $\sigma=4$, $w=0.9$ and we split 7129 entries into 7 batches.

\begin{table}[H]
\centering
\begin{tabular}{ccc}
\hline
Method & Training error & Test error   \\
\hline
FAIR & 1/38 & 1/34 \\
EB & 0/38 & 3/34 \\
IR & 1/38 & 6/34\\
DP & 1/38 & 2/34\\
Sparse DP & 1/38 & 2/34\\
Hard Thresh DP & 1/38 & 2/34\\
\hline
\end{tabular}
\caption{Training error and test error of leukemia data set}
\end{table}

From Table 7 we conclude DP family classifiers outperform EB classifier. EB classifier has the same performance as IR. Improvement of EB compared with IR is marginal but using DP and Sparse DP classifier could result in some improvement. Both DP and EB classifier shrink the mean difference but doesn't eliminate any irrelevant feature. Sparse DP classifier selects 2092 features but has the same performance in terms of test error as DP classifier. This might be due to the fact that this dataset is relatively well separated. Thresholding might not improve a lot.

\section{Discussion}
The contribution of this paper is three-folds: first we established the relationship between the estimation error and the classification error theoretically; second we proposed two empirical Bayes estimators for the normalized mean difference and the induced linear classifiers. Third, for estimating $\boldsymbol\eta$, we develop a variational Bayes algorithm to approximate posterior distribution of Dirichlet process mixture model with a special base measure and we could parallelize our algorithm using the ``batch'' idea.

Yet, there are still many open problems and many possible extensions related to this work. For example, instead of using the Independence Rule, we could develop a Bayes procedure to threshold both the mean difference and the sample covariance matrix, in a spirit similar to  \citet{shao2011sparse}, \citet{cai2012direct} and \citet{bickel2008regularized}.
%we could use the thresholded sample covariance matrix estimator proposed in \citet{bickel2008regularized} since it's a consistent estimator when $\log p=o(n)$.
Another extension is to relax normality assumption: LDA is suitable for any elliptical distribution, therefore our work could also be extended to a bigger family of distributions under sub-Gaussian constraints.

%{\noindent \em Remainder omitted in this sample. See http://www.jmlr.org/papers/ for full paper.}
%
%% Acknowledgements should go at the end, before appendices and references
%
%\acks{We would like to acknowledge support for this project
%from the National Science Foundation (NSF grant IIS-9988642)
%and the Multidisciplinary Research Program of the Department
%of Defense (MURI N00014-00-1-0637). }


%\newpage

\appendix
\section{Proofs of 2 Theorems}
\begin{proof}[Proof of Theorem 1]
$\Psi$ could be written as
\[\frac{(\boldsymbol{\mu}_1-\hat{\boldsymbol{\mu}})^t\hat{\mathbf{D}}^{-\frac{1}{2}}\hat{\boldsymbol{\eta}}}{\sqrt{\hat{\boldsymbol{\eta}}^t\hat{\mathbf{D}}^{-\frac{1}{2}}\Sigma \hat{\mathbf{D}}^{-\frac{1}{2}}\hat{\boldsymbol{\eta}}}},\]
which is lower bounded by
\[\tilde{\Psi}=\frac{(\boldsymbol{\mu}_1-\hat{\boldsymbol{\mu}})^t\hat{\mathbf{D}}^{-\frac{1}{2}}\hat{\boldsymbol{\eta}}}{\sqrt{\lambda_1\hat{\boldsymbol{\eta}}^t\hat{\mathbf{D}}^{-\frac{1}{2}}\mathbf{D}\hat{\mathbf{D}}^{-\frac{1}{2}}\hat{\boldsymbol{\eta}}}}.\]
By Lemma A.2. of \cite{fan2008high} we have $\max_{i\leq p}|\hat\sigma_{ii}-\sigma_{ii}|\rightarrow_p 0$. Therefore $\hat{\mathbf{D}}=\mathbf{D}(1+o_p(1))$. Therefore we have $\tilde{\Psi}=\frac{({\boldsymbol{\mu}}_1-\hat{{\boldsymbol{\mu}}})\mathbf{D}^{-\frac{1}{2}}\hat{\boldsymbol{\eta}}}{\sqrt{\lambda_1}\|\hat{\boldsymbol{\eta}}\|}(1+o_p(1))$.

We first consider the denominator,
\[\|\hat{\boldsymbol{\eta}}\|=\sqrt{\sum^{p}_{i=1}\hat\eta_i^2}=\sqrt{\sum_{i\in S}\hat\eta_i^2+\sum_{i\in S^c}\hat\eta_i^2}=\sqrt{\|\hat{\boldsymbol{\eta}}_1\|^2+\|\hat{\boldsymbol{\eta}}_2\|^2};\]

If $\eta_i$=0, $Y_i\sim t_{2n-2}$, according to $t$ distribution tail probability inequality, we have
\[P(|Y_i|\geq b_n)\leq \left(\frac{2n-2}{2n-3}\cdot \frac{\Gamma(\frac{2n-1}{2})}{\sqrt{\pi (2n-2)}\Gamma(n-1)}\right)\cdot \frac{1}{b_n}(1+\frac{b_n^2}{2n-2})^{-\frac{2n-1}{2}}.\]

for any $\epsilon>0$, using Markov Inequality and Cauchy-Schwartz Inequality,
\begin{align*}
P(\|\hat{\boldsymbol{\eta}}_2\|^2>\epsilon)&\leq \frac{1}{\epsilon}\sum_{i\in S^c}\mathds{E}[\hat\eta_i^21_{\{|Y_i|\geq b_n\}}] \leq \frac{1}{\epsilon}\sum_{i \in S^c}\sqrt{\mathds{E}(\hat\eta_i^4)}\sqrt{P(|Y_i|\geq b_n)}\\
&\leq \frac{\sup_{i \in S^c}\sqrt{\mathds{E}(\hat\eta_i^4)}}{\epsilon}\sum_{i \in S^c}\sqrt{P(|Y_i|\geq b_n)}\\
&\leq \frac{\sup_{i \in S^c}\sqrt{\mathds{E}(\hat\eta_i^4)}}{\epsilon}(p-s_n) \sqrt{\frac{2n-2}{2n-3}\cdot \frac{2\Gamma(\frac{2n-1}{2})}{\sqrt{\pi (2n-2)}\Gamma(n-1)}}\\
&\cdot \sqrt{\frac{1}{b_n}(1+\frac{b_n^2}{2n-2})^{-\frac{2n-1}{2}}}\\
&\sim \sqrt[4]{\frac{2}{\pi}}\frac{\sup_{i \in S^c}\sqrt{\mathds{E}(\hat\eta_i^4)}}{\epsilon}
(p-s_n)\cdot \sqrt{\frac{1}{b_n}(1+\frac{b_n^2}{2n-2})^{-\frac{2n-1}{2}}}\\
&\sim \sqrt[4]{\frac{2}{\pi}}\frac{\sup_{i \in S^c}\sqrt{\mathds{E}(\hat\eta_i^4)}}{\epsilon} \exp(\log(p-s_n)-\frac{1}{2}\log(b_n)-\frac{2n-1}{2n-2}\cdot \frac{b_n^2}{4})\rightarrow 0;
\end{align*}
The last equivalence holds since $b_n^2/n\rightarrow 0$. The probability goes to 0 since $\log (p-s_n)=o(b^2_n)$.

Remember $C_p= \mathbf{d}^t\mathbf{D}^{-1}\mathbf{d}= (2/n)\|\boldsymbol{\eta}_1\|^2$.
According to triangular inequality,
\begin{align*}
\|\hat{\boldsymbol{\eta}}_1\|&\leq \|\hat{\boldsymbol{\eta}}_1-\boldsymbol{\eta}_1\|+\|\boldsymbol{\eta}_1\|\\
&=O_p(\sqrt{\varepsilon_n})+\|\boldsymbol{\eta}_1\|.
\end{align*}

We put these terms together to approximate the order of denominator as $O_p(\sqrt{\varepsilon_n})+\|\boldsymbol{\eta}_1\|$.

For numerator, denote ${\boldsymbol{\mu}}_1=({{\mu}}_{11},{{\mu}}_{12}, \cdots, {{\mu}}_{1p})^t$, ${\boldsymbol{\mu}}_2=({{\mu}}_{21},{{\mu}}_{22}, \cdots, {{\mu}}_{2p})^t$ and $\hat{{\boldsymbol{\mu}}}=(\hat{{{\mu}}}_{1},\hat{{{\mu}}}_{2}, \cdots, \hat{{{\mu}}}_{p})^t$, where $\hat{{{\mu}}}_i=(\hat{{\mu}}_{1i}+\hat{{\mu}}_{2i})/2$. we have the following decomposition:
\[({\boldsymbol{\mu}}_1-\hat{{\boldsymbol{\mu}}})^t\mathbf{D}^{-\frac{1}{2}}\hat{\boldsymbol{\eta}}=\sum^p_{i=1}({{\mu}}_{1i}-\hat{{{\mu}}}_{i})\sigma^{-\frac{1}{2}}_{ii}\hat\eta_i=\sum_{i \in S^c}({{\mu}}_{1i}-\hat{{{\mu}}}_{i})\sigma^{-\frac{1}{2}}_{ii}\hat\eta_i+\sum_{i\in S}({{\mu}}_{1i}-\hat{{{\mu}}}_{i})\sigma^{-\frac{1}{2}}_{ii}\hat\eta_i\equiv I_1+I_2.\]
For $I_1$, we have the following decomposition since ${\boldsymbol{\mu}}_{1i}={\boldsymbol{\mu}}_{2i}$:
\[I_1=-\frac{1}{2}\sum_{i \in S^c}(\hat{{\boldsymbol{\mu}}}_{1i}-{\boldsymbol{\mu}}_{1i})\sigma^{-\frac{1}{2}}_{ii}\hat\eta_i-\frac{1}{2}\sum_{i \in S^c}(\hat{{\boldsymbol{\mu}}}_{2i}-{\boldsymbol{\mu}}_{2i})\sigma^{-\frac{1}{2}}_{ii}\hat\eta_i\equiv\frac{1}{2}I_{1,1}+\frac{1}{2}I_{1,2}.\]
Using Markov Inequality,
\begin{align*}
P(|I_{1,1}|>\epsilon)&\leq \frac{1}{\epsilon}\sum_{i \in S^c}\mathds{E}|(\hat{{{\mu}}}_{1i}-{{\mu}}_{1i})\sigma^{-\frac{1}{2}}_{ii}\hat\eta_i1_{\{|Y_i|\geq b_n\}}|\\
& \leq \frac{k^{-1/2}_1}{\epsilon}\sum_{i \in S^c}\sqrt{\mathds{E}(\hat{{{\mu}}}_{1i}-{{\mu}}_{1i})^2}\sqrt{E(\hat\eta^2_i1_{\{|Y_i|\geq b_n\}})}\\
&  \leq \frac{k^{-1/2}_1k^{1/2}_2}{\epsilon}\sum_{i \in S^c} \frac{1}{\sqrt{n}} \sqrt[4]{\mathds{E}(\hat\eta_i^4)}\sqrt[4]{P(|Y_i|\geq b_n)}\\
& \sim \frac{k^{-1/2}_1k^{1/2}_2\sqrt[8]{\frac{2}{\pi}}\sup_{i \in S^c}\sqrt[4]{\mathds{E}(\hat\eta_i^4)}}{\sqrt{n}\epsilon}\exp(\frac{\log(p-s_n)}{2}-\frac{\log(b_n)}{4}-\frac{2n-1}{2n-2}\cdot \frac{b_n^2}{8}) \rightarrow 0.
\end{align*}
Therefore $I_{1,1}=o_p(1)$. Similarly $I_{1,2}=o_p(1)$. Hence $I_1=o_p(1)$. Suppose $\mathbf{D}=\text{diag}(\text{diag}(\mathbf{D}_1),\text{diag}(\mathbf{D}_2))$ where $\mathbf{D}_1$ denotes the corresponding submatrix of relevant features and $\mathbf{D}_2$ denotes the corresponding submatrix of irrelevant features. Similarly for $\mathbf{R}$ denote the corresponding submatrix of irrelevant features as $\mathbf{R}_1$.
Denote the sub-vector of ${\boldsymbol{\mu}}_1=({\boldsymbol{\mu}}^\ast_1,\mathbf{0}^t_{p-s_n})^t$ and ${\boldsymbol{\mu}}_2=(({\boldsymbol{\mu}}_2^\ast)^t,\mathbf{0}^t_{p-s_n})^t$.  For $I_2$ we have the following decomposition:

\begin{align*}
I_2&=\sqrt{\frac{n}{8}}({\boldsymbol{\mu}}_1^\ast-{\boldsymbol{\mu}}_2^\ast)^t\mathbf{D}_1^{-1}({\boldsymbol{\mu}}_1^\ast-{\boldsymbol{\mu}}_2^\ast)-\frac{1}{2}(\hat{{\boldsymbol{\mu}}}^\ast_1-{\boldsymbol{\mu}}^\ast_1)^t\mathbf{D}_1^{-\frac{1}{2}}(\hat{\boldsymbol{\eta}}_1-\boldsymbol{\eta}_1)\\
&-\frac{1}{2}(\hat{{\boldsymbol{\mu}}}^\ast_2-{\boldsymbol{\mu}}^\ast_2)^t\mathbf{D}_1^{-\frac{1}{2}}(\hat{\boldsymbol{\eta}}_1-\boldsymbol{\eta}_1)-\frac{1}{2}(\hat{{\boldsymbol{\mu}}}^\ast_1-{\boldsymbol{\mu}}^\ast_1)^t\mathbf{D}_1^{-\frac{1}{2}}\boldsymbol{\eta}_1-\frac{1}{2}(\hat{{\boldsymbol{\mu}}}^\ast_2-{\boldsymbol{\mu}}^\ast_2)^t\mathbf{D}_1^{-\frac{1}{2}}\boldsymbol{\eta}_1\\
&+(\frac{{\boldsymbol{\mu}}^\ast_1-{\boldsymbol{\mu}}^\ast_2}{2})^t\mathbf{D}_1^{-\frac{1}{2}}(\hat{\boldsymbol{\eta}}_1-\boldsymbol{\eta}_1)\\
&\equiv \sqrt{\frac{1}{2n}}\|\boldsymbol{\eta}_1\|^2-\frac{1}{2}I_{2,1}-\frac{1}{2}I_{2,2}-\frac{1}{2}I_{2,3}-\frac{1}{2}I_{2,4}+\frac{1}{2}I_{2,5}.
\end{align*}

Since $\hat{{\boldsymbol{\mu}}}^\ast_1-{\boldsymbol{\mu}}^\ast_1\sim N(\mathbf{0},\frac{1}{n}\mathbf{D}_1^{-1/2}\mathbf{R}_1\mathbf{D}_1^{-1/2})$, we use Cauchy-Schwartz Inequality to get an upper bound. We have
\[I^2_{2,1}=\frac{1}{4}((\hat{{\boldsymbol{\mu}}}^\ast_1-{\boldsymbol{\mu}}^\ast_1)^t\mathbf{D}_1^{-\frac{1}{2}}(\hat{\boldsymbol{\eta}}_1-\boldsymbol{\eta}_1))^2 \leq\frac{1}{4}(\hat{{\boldsymbol{\mu}}}^\ast_1-{\boldsymbol{\mu}}^\ast_1)^t\mathbf{D}_1^{-1}(\hat{{\boldsymbol{\mu}}}^\ast_1-{\boldsymbol{\mu}}^\ast_1) \cdot \|\hat{\boldsymbol{\eta}}_1-\boldsymbol{\eta}_1\|^2.\]

 $(\hat{{\boldsymbol{\mu}}}^\ast_1-{\boldsymbol{\mu}}^\ast_1)^t\mathbf{D}_1^{-1}(\hat{{\boldsymbol{\mu}}}^\ast_1-{\boldsymbol{\mu}}^\ast_1)$ is $O_p(\frac{s_n}{n}\lambda_{\max}(\mathbf{R}_1))=O_p(\frac{s_n}{n})$, meanwhile, $\|\hat{\boldsymbol{\eta}}_1-\boldsymbol{\eta}_1\|^2$ is $O_p(\varepsilon_n)$. Therefore $I_{2,1}=O_p(\sqrt{\frac{s_n\varepsilon_n}{n}})$.
Similarly $I_{2,2}=O_p(\sqrt{\frac{s_n\varepsilon_n}{n}})$.

Note that $I_{2,3}\sim N(0,\frac{1}{4n}\boldsymbol{\eta}^t_1\mathbf{R}_1\boldsymbol{\eta}_1)$. $\lambda_{\max}(\mathbf{R}_1)\leq \lambda_1$, therefore $I_{2,3}=O_p(\|\boldsymbol{\eta}_1\|/\sqrt{n})$. Similarly $I_{2,4}=O_p(\|\boldsymbol{\eta}_1\|/\sqrt{n})$.

For $I_{2,5}$, according to Cauchy-Schwartz Inequality, we have

\[|I_{2,5}| \leq \frac{1}{2}\sqrt{\frac{2}{n}\|\boldsymbol{\eta}_1\|^2\cdot \|\hat{\boldsymbol{\eta}}_1-\boldsymbol{\eta}_1\|^2}=O_p(\sqrt{\varepsilon_n/n}\|\boldsymbol{\eta}_1\|).\]

Asymptotically, we have
\[\tilde\Psi=\frac{\sqrt{\frac{1}{2n}}\|\boldsymbol{\eta}_1\|^2-O_p(\sqrt{\frac{s_n\varepsilon_n}{n}})-O_p(\sqrt{\varepsilon_n/n}\|\boldsymbol{\eta}_1\|)-O_p(\|\boldsymbol{\eta}_1\|/\sqrt{n})}{\sqrt{\lambda_1}(O_p(\sqrt{\varepsilon_n})+\|\boldsymbol{\eta}_1\|)}(1+o_p(1)).\]
Therefore
\begin{footnotesize}
\begin{equation}\label{eq:prederror}
W(\hat{\delta}_{\hat{\boldsymbol\theta}},\boldsymbol{\theta})\leq 1-\Phi\left(\frac{\sqrt{\frac{1}{2n}}\|\boldsymbol{\eta}_1\|^2-O_p(\sqrt{\frac{s_n\varepsilon_n}{n}})-O_p(\sqrt{\varepsilon_n/n}\|\boldsymbol{\eta}_1\|)-O_p(\|\boldsymbol{\eta}_1\|/\sqrt{n})}{\sqrt{\lambda_1}(O_p(\sqrt{\varepsilon_n})+\|\boldsymbol{\eta}_1\|)}(1+o_p(1))\right).
\end{equation}
\end{footnotesize}
Since $\|\boldsymbol{\eta}_1\|^2= nC_p/2$, we have
\[W(\hat{\delta}_{\hat{\boldsymbol\theta}})\leq 1-\Phi\left(\frac{\sqrt{n/8}C_p-O_p(\sqrt{\frac{s_n\varepsilon_n}{n}})-O_p(\sqrt{\varepsilon_nC_p})-O_p(\sqrt{C_p})}{\sqrt{\lambda_1}(\sqrt{nC_p/2}+O_p(\sqrt{\varepsilon_n}))}(1+o_p(1))\right).\]

If $\frac{\varepsilon_n}{n}=o(C_p)$, $\frac{\sqrt{s_n\varepsilon_n}}{n}=o(C_p)$ and $nC_p\rightarrow \infty$, then $\sqrt{n/8}C_p$ and $\sqrt{\lambda_1}\sqrt{nC_p/2}$ are the leading terms of denominator and numerator respectively. We have
\[W(\hat{\delta})\leq 1-\Phi\left(\frac{\sqrt{C_p}}{2\sqrt{\lambda_1}}(1+o_p(1))\right).\]

\end{proof}
\begin{proof}[Proof of Theorem 2]
Conditions in Theorem 2 implies the conditions in Theorem 1. Therefore
\[W(\hat{\delta}_{\hat{\boldsymbol\theta}})\leq\Phi\left(-\frac{\sqrt{n/8}C_p-O_p(\sqrt{\frac{s_n\varepsilon_n}{n}})-O_p(\sqrt{\varepsilon_nC_p})-O_p(\sqrt{C_p})}{\sqrt{\lambda_1}(\sqrt{nC_p/2}+O_p(\sqrt{\varepsilon_n}))}(1+o_p(1))\right).\]
Using Lemma 1 in \cite{shao2011sparse}, we let $\xi_n=\frac{C_p}{4\lambda_1}$ and
\[\tau_n=\frac{O_p(\sqrt{\frac{s_n\varepsilon_n}{n}})+O_p(\sqrt{\varepsilon_nC_p})+O_p(\sqrt{C_p})}{O_p(\sqrt{\varepsilon_nC_p})+\sqrt{\lambda_1/2}\sqrt{n}C_p}.\]
Using the conditions $\varepsilon_nC_p=o(n)$ and $\sqrt{s_n\varepsilon_n}=o(n)$, we could easily verify that $\xi_n\rightarrow \infty$, $\tau_n\rightarrow 0$ and $\tau_n\xi_n\rightarrow 0$, therefore
\[W(\hat{\delta}_{\hat{\boldsymbol\theta}})/W(\delta_{OPT})\rightarrow 0.\]
\end{proof}

\section{Variational Inference Algorithm Derivation}
We will derive the variational inference algorithm for Dirichlet process mixture model. $\alpha$, $T$, $w$, $\sigma^2$ and the data vector $\mathbf{y}$ is given in advance. The data generating process is summarized in (3)-(8). We treat $\mathbf{Z}$ as latent variables and $\mathbf{V},\boldsymbol{\eta^\ast},\boldsymbol{\xi}$ as parameters. Posterior distribution of all the parameters and latent variables is proportional to
\begin{align*}
P(\mathbf{Z},\mathbf{V},\boldsymbol{\eta^\ast},\boldsymbol{\xi}|\mathbf{y})& \propto P(\mathbf{y},\mathbf{Z},\mathbf{V},\boldsymbol{\eta^\ast},\boldsymbol{\xi})=P(\boldsymbol{\xi}|w)P(\mathbf{V}|\boldsymbol{\alpha})P(\boldsymbol{\eta^\ast}|\boldsymbol{\xi})P(\mathbf{Z}|\mathbf{V})P(\mathbf{y}|\mathbf{Z},\boldsymbol{\eta^\ast})\\
&\propto w^{\sum^T_{t=1}\xi_t}(1-w)^{T-\sum^T_{t=1}\xi_t}\prod^{T-1}_{t=1}(1-V_t)^{\alpha-1}\prod_{t:\xi_t=1}\delta_0(\eta^\ast_t)\cdot\\
&\prod_{t:\xi_t=0}\frac{1}{\sqrt{2\pi}\sigma}\exp(-\frac{(\eta^\ast_t)^2}{2\sigma^2})\cdot\prod_{t=1}^T\pi_t^{\sum^p_{k=1}1_{Z_k=t}}\cdot\\
&\exp\left(-\frac{\sum^p_{k=1}\sum^T_{t=1}(Y_k-\eta^\ast_t)^21_{Z_k=t}}{2}\right).
\end{align*}
Recall that under the fully factorized variational assumption, we have
\[q(\mathbf{Z},\mathbf{V},\boldsymbol{\eta^\ast},\boldsymbol{\xi})=q_{\mathbf{p},\mathbf{m},\boldsymbol{\tau}}(\boldsymbol{\eta^\ast},\boldsymbol{\xi})q_{\boldsymbol{\gamma}_1,\boldsymbol{\gamma}_2}(\mathbf{V})q_{\boldsymbol{\Phi}}(\mathbf{Z}).\]
Define $P(Z_k=t)=\phi_{k,t}$. First we find the optimal form of $q(\boldsymbol{\eta^\ast},\boldsymbol{\xi})$, which satisfies
\begin{align*}
\log q(\boldsymbol{\eta^\ast},\boldsymbol{\xi})&=\mathds{E}_{\mathbf{V},\mathbf{Z}}[\log(w^{\sum^T_{t=1}\xi_t}(1-w)^{T-\sum^T_{t=1}\xi_t}\prod_{t:\xi_t=1}\delta_0(\eta^\ast_t)\cdot\\
&\prod_{t:\xi_t=0}\frac{1}{\sqrt{2\pi}\sigma}\exp(-\frac{(\eta^\ast_t)^2}{2\sigma^2})\exp(-\frac{\sum^p_{k=1}\sum^T_{t=1}(Y_k-\eta^\ast_t)^21_{Z_k=t}}{2}))]+\text{const}\\
&=\sum^T_{t=1}[1_{\xi_t=1}(\log w+\log \delta_0(\eta^\ast_t))+1_{\xi_t=0}(\log (1-w)-\log \sqrt{2\pi \sigma^2}-\frac{(\eta^\ast_t)^2}{2\sigma^2})\\
&-\frac{\sum^p_{k=1}\phi_{k,t}(Y_k-\eta^\ast_t)^2}{2}]+\text{const}\\
&=\sum^T_{t=1}[1_{\xi_t=1}(\log w+\log \delta_0(\eta^\ast_t)-\frac{\sum^p_{k=1}\phi_{k,t}Y_k^2}{2})\\
&+1_{\xi_t=0}(\log (1-w)-\log \sqrt{2\pi \sigma^2}-\frac{(\eta^\ast_t)^2}{2\sigma^2}-\frac{\sum^p_{k=1}\phi_{k,t}(Y_k-\eta^\ast_t)^2}{2})+\text{const}]\\
&\equiv \sum^T_{t=1}\log q(\xi_t,\eta^\ast_t);
\end{align*}
where $\log q(\xi_t,\eta^\ast_t)=1_{\xi_t=1}(\log w+\log \delta_0(\eta^\ast_t)-\frac{\sum^p_{k=1}\phi_{k,t}Y_k^2}{2})
+1_{\xi_t=0}(\log (1-w)-\log \sqrt{2\pi \sigma^2}-\frac{(\eta^\ast_t)^2}{2\sigma^2}-\frac{\sum^p_{k=1}\phi_{k,t}(Y_k-\eta^\ast_t)^2}{2})+\text{const}$. Therefore the optimal form of $q_{\mathbf{p},\mathbf{m},\boldsymbol{\tau}}(\boldsymbol{\eta^\ast},\boldsymbol{\xi})$ is fully factorized across different clusters:
$q_{\mathbf{p},\mathbf{m},\boldsymbol{\tau}}(\boldsymbol{\eta^\ast},\boldsymbol{\xi})=\prod^T_{t=1}q_{p_t,m_t,\tau_t}(\eta^\ast_t,\xi_t).$
In order to determine the updating formula for $p_t,m_t,\tau_t$, we use Method of Undetermined Coefficients. Suppose $q(\xi_t,\eta^\ast_t)= p_t1_{\xi_t=1}\delta_0(\eta^\ast_t)+(1-p_t)1_{\xi_t=0} (2\pi\sigma^2_t)^{-1/2}\exp(-(\eta^\ast_t-m_t)^2/(2\sigma^2_t))$, therefore $\log q(\xi_t,\eta^\ast_t)=1_{\xi_t=1}(\log p_t+\log(\delta_0(\eta^\ast_t)))+1_{\xi_t=0}(\log(1-p_t)-\log(\sqrt{2\pi \tau_t^2})-\frac{(\eta^\ast_t-m_t)^2}{2\tau_t^2})+\text{const}$. Even though there's a normalizing constant, but the difference between multipliers of $1_{\xi_t=1}$ and $1_{\xi_t=0}$ is invariant with respect to the constant. Therefore we have the following equation:
\begin{align*}
&\log p_t-\log(1-p_t)+\log \sqrt{2\pi \tau_t^2}+\frac{(\eta^\ast_t-m_t)^2}{2\tau_t^2}=\\
&\log w-\log(1-w)-\frac{\sum^p_{k=1}\phi_{k,t}Y_k^2}{2}+\frac{(\eta^\ast_t)^2}{2\sigma^2}+\frac{\sum^p_{k=1}\phi_{k,t}(Y_k-\eta^\ast_t)^2}{2};
\end{align*}
which holds for any $\eta^\ast_t \in \mathds{R}$. The solutions are given as follows:
\begin{align*}
m_t&=\frac{\sigma^2\cdot \sum^p_{k=1}\phi_{k,t}Y_k}{\sigma^2\cdot \sum^p_{k=1}\phi_{k,t}+1}, t=1,2,\cdots, T\\
\tau_t^2&=\frac{\sigma^2}{\sigma^2\cdot \sum^p_{k=1}\phi_{k,t}+1}, t=1,2,\cdots,T\\
p_t&=\frac{\exp\left(\log(w)-\log(1-w)+\log(\sqrt{\sigma^2\cdot \sum^p_{k=1}\phi_{k,t}+1})-\frac{\sigma^2\cdot (\sum^p_{k=1}\phi_{k,t}Y_k)^2}{2(\sigma^2\cdot \sum^p_{k=1}\phi_{k,t}+1)}\right)}{\exp\left(\log(w)-\log(1-w)+\log(\sqrt{\sigma^2\cdot \sum^p_{k=1}\phi_{k,t}+1})-\frac{\sigma^2\cdot (\sum^p_{k=1}\phi_{k,t}Y_k)^2}{2(\sigma^2\cdot \sum^p_{k=1}\phi_{k,t}+1)}\right)+1},\\
t&=1,2,\cdots,T.
\end{align*}

Next we deal with the optimal form for $q(\mathbf{V})$, which satisfies
\begin{align*}
\log q(\mathbf{V})&=\mathds{E}_\mathbf{Z}[\log(\prod^{T-1}_{t=1}(1-V_t)^{\alpha-1}\cdot V_1^{\sum^p_{k=1}1_{Z_k=1}}\cdot (V_2(1-V_1))^{\sum^p_{k=1}1_{Z_k=2}}\cdots\\ &(V_{T-1}\prod_{t=1}^{T-2}(1-V_t))^{\sum^p_{k=1}1_{Z_k=T-1}}(\prod^{T-1}_{t=1}(1-V_{t}))^{\sum^p_{k=1}1_{Z_k=T}})]+\text{const}\\
&=\sum^p_{k=1}\phi_{k,1}\cdot \log V_1+(\alpha-1+\sum^T_{t=2}\sum^p_{k=1}\phi_{k,t})\log (1-V_1)+\sum^p_{k=1}\phi_{k,2}\log V_2+(\alpha-1+\\
&\sum^T_{t=3}\sum^p_{k=1}\phi_{k,t})\cdot \log(1-V_2)+\cdots+\sum^{p}_{k=1}\phi_{k,T-1}\cdot \log V_{T-1}\\
&+(\alpha-1+\sum^p_{k=1}\phi_{k,T})\log(1-V_{T-1})+\text{const}\\
&\equiv \sum^{T-1}_{t=1}\log q(V_t);
\end{align*}

where $\log q(V_1)=\sum^p_{k=1}\phi_{k,1}\cdot \log V_1+(\alpha-1+\sum^T_{t=2}\sum^p_{k=1}\phi_{k,t})\log (1-V_1)+\text{const}$, $\log q(V_2)=\sum^p_{k=1}\phi_{k,2}\log V_2+(\alpha-1+\sum^T_{t=3}\sum^p_{k=1}\phi_{k,t})\cdot \log(1-V_2)+\text{const}$, $\cdots$, $\log q(V_{T-1})=\sum^{p}_{k=1}\phi_{k,T-1}\cdot \log V_{T-1}+(\alpha-1+\sum^p_{k=1}\phi_{n,T})\log(1-V_{T-1})+\text{const}$. Thus we proved
\[q_{\boldsymbol{\gamma}_1,\boldsymbol{\gamma}_2}(\mathbf{V})=\prod^{T-1}_{t=1}q_{\gamma_{1t},\gamma_{2t}}(V_t).\]
Besides, $V_1$ follows Beta Distribution with parameters $(\gamma_{11},\gamma_{21})=(\sum^p_{k=1}\phi_{k,1}+1,\alpha+\sum^T_{t=2}\sum^p_{k=1}\phi_{k,t})$, $V_2$ follows Beta Distribution with parameters $(\gamma_{12},\gamma_{22})=(\sum^p_{k=1}\phi_{k,2}+1,\alpha+\sum^T_{t=3}\sum^p_{k=1}\phi_{k,t})$,$\cdots$, $V_{T-1}$ follows Beta Distribution with parameters $(\gamma_{T-1,2},\gamma_{T-1,2})=(\sum^p_{k=1}\phi_{k,T-1}+1,\alpha+\sum^p_{k=1}\phi_{k,T})$.

Finally, we deal with $q(\mathbf{Z})$. We have the following optimal form
\begin{align*}
\log q(\mathbf{Z})&=\mathds{E}_{\boldsymbol{\xi},\boldsymbol{\eta^\ast},\mathbf{V}}[\log(\prod^T_{t=1}\pi_t^{\sum^p_{k=1}1_{Z_k=t}})\cdot\exp(-\frac{\sum^p_{k=1}\sum^T_{t=1}(Y_k-\eta^\ast_t)^21_{Z_k=t}}{2})]+\text{const}\\
&=\sum^p_{k=1}\mathds{E}_{\boldsymbol{\xi},\boldsymbol{\eta^\ast},\mathbf{V}}[\sum^T_{t=1}(\log \pi_t-\frac{(Y_k-\eta^\ast_t)^2}{2})1_{Z_k=t}]+\text{const}\\
&=\sum^p_{k=1}\sum^T_{t=1}(\mathds{E}_{\mathbf{V}}(\log \pi_t)-\frac{\mathds{E}_{\xi_t,\eta^\ast_t}(Y_k-\eta^\ast_t)^2}{2})1_{Z_k=t}+\text{const}\\
&=\sum^p_{k=1}\log q(Z_k);
\end{align*}
therefore we proved $q_{\boldsymbol{\Phi}}(\mathbf{Z})=\prod^N_{n=1}q_{\boldsymbol{\phi}_n}(Z_k)$.
Since $\mathds{E}_{\xi_t,\eta^\ast_t}(Y_k-\eta^\ast_t)^2=Y_k^2-2(1-p_t)m_tY_k+(1-p_t)(m_t^2+\tau^2_t)$, we have
\begin{align*}
\log q(Z_k)&=\sum^T_{t=1}(\mathds{E}_{\mathbf{V}}(\log \pi_t)+(1-p_t)m_tY_k-\frac{1}{2}(1-p_t)(m_t^2+\tau_t^2))1_{Z_k=t}+\text{const}\\
&=\sum^T_{t=1}[\mathds{E}_{\gamma_{1,t},\gamma_{2,t}}(\log V_t)+\sum^{t-1}_{i=1}\mathds{E}_{\gamma_{1,i},\gamma_{2,i}}(\log (1-V_i))\\
&+(1-p_t)m_tY_k-\frac{1}{2}(1-p_t)(m_t^2+\tau_t^2)]1_{Z_k=t}+\text{const}.
\end{align*}

Therefore $q(Z_k)$ is the probability mass function of Multinomial Distribution. Once we fix $k$, $\phi_{k,t}\propto \exp(S_t)$, where $S_t=\exp[\mathds{E}_{\gamma_{1,t},\gamma_{2,t}}(\log V_t)+\sum^{t-1}_{i=1}\mathds{E}_{\gamma_{1,i},\gamma_{2,i}}(\log( 1-V_i))+(1-p_t)m_tY_k-\frac{1}{2}(1-p_t)(m_t^2+\tau_t^2)]$.

\bibliography{classificationsample}

\end{document}